# Feature Selection Based on Reinforcement Learning and Hazard State Classification for Magnetic Adhesion Wall-Climbing Robots

Zhen Ma[a], He Xu[a*], Jielong Dou[a], Yi Qin[a], Xueyu Zhang[a]

[a]College of Mechanical and Electrical Engineering, Harbin Engineering University, Harbin 150001, China. No.145, Nan Tong Street, Nan Gang District.

**Abstract:** Magnetic adhesion tracked wall-climbing robots face potential risks of overturning during high-altitude operations, making their stability crucial for ensuring safety. This study presents a dynamic feature selection method based on Proximal Policy Optimization (PPO) reinforcement learning, combined with typical machine learning models, aimed at improving the classification accuracy of hazardous states under complex operating conditions. Firstly, this work innovatively employs a fiber rod-based MEMS attitude sensor to collect vibration data from the robot and extract high-dimensional feature vectors in both time and frequency domains. Then, a reinforcement learning model is used to dynamically select the optimal feature subset, reducing feature redundancy and enhancing classification accuracy. Finally, a CNN-LSTM deep learning model is employed for classification and recognition. Experimental results demonstrate that the proposed method significantly improves the robot's ability to assess hazardous states across various operational scenarios, providing reliable technical support for robotic safety monitoring.

## 1. Introduction

Magnetic adhesion tracked wall-climbing robots are designed specifically for vertical or inclined surfaces, enabling them to effectively counteract gravity and perform a variety of tasks [1], such as inspection, welding, and cleaning in high-altitude environments [2-5]. These robots have broad application prospects, particularly in dangerous high-altitude operations, where they can significantly improve work efficiency and ensure the safety of operators [6]. However, as the robot moves along the wall, the overturning torque generated by its weight and load may cause it to flip backward, affecting its stability and posing potential safety risks [7]. Therefore, ensuring the robot's stability during operation, especially by maintaining sufficient magnetic attachment force and continuously monitoring the attachment state of the magnetic pads, is crucial for its safety [8-9].

To address this issue effectively, developing a technology capable of real-time sensing and evaluating the robot's magnetic attachment status is essential [10]. In this regard, Micro-Electromechanical Systems (MEMS) attitude sensors offer an ideal solution. These sensors can monitor the robot's angular velocity, acceleration, and magnetic field strength in real time and have played a significant role in robot balance control, human motion analysis, and aircraft attitude measurement [11-13]. By collecting data from these sensors, it becomes possible to analyze and assess the wall-climbing robot's motion state during operation, providing a basis for identifying hazardous states [14].

Although existing data analysis methods based on MEMS sensors can identify the robot's hazardous states to some extent [5], the data from MEMS sensors often requires feature extraction in both time and frequency domains to be effectively utilized [16]. Due to the high dimensionality of the feature space, many features may exhibit redundancy or similarity, which poses challenges for classification accuracy and computational efficiency [17]. To address this issue, identifying the most representative features and reducing feature redundancy in the high-dimensional feature space are key to improving classification performance.

Traditional feature selection methods are typically classified into three categories: filter methods, wrapper methods, and embedded methods. Filter methods select features independently of the learning algorithm, while wrapper methods evaluate the predictive performance of feature subsets to make selections. Embedded methods, such as decision trees, automatically perform feature selection during the training process [18-19]. However, due to significant differences in the robot's motion characteristics across different operational states [20], traditional feature selection methods often overlook the complex relationships between features. Therefore, a feature selection method that accounts for the dynamic relationships within the data is necessary. Such a method would allow for dynamic selection of the optimal feature subset based on the robot's current state and environmental conditions, without relying on a specific dataset from the robot's operating conditions.

In the reinforcement learning framework, feature selection is regarded as a typical decision-making process [21]. In this framework, each feature subset is defined as a "state," and the action of feature selection is performed by choosing a new feature and adding it to the current feature subset [22-23]. Based on previous research, to optimize the feature selection process, we have designed a multi-level reward function that evaluates the effectiveness of selecting a feature subset based on the wall-climbing robot's current task requirements and environmental changes. Through the interaction between the reinforcement learning algorithm and the environment, the robot can gradually adjust its feature selection strategy, adapting to different data distributions and obtaining the optimal feature subset under various operational conditions.

Therefore, this paper proposes a PPO-based reinforcement learning feature selection model to select effective features from the high-dimensional feature data of MEMS attitude sensors. By using a typical CNN-LSTM deep learning classification model, the robot's hazardous states are evaluated and classified effectively. The proposed research framework is shown in Fig. 1.

## 2. Literature Review

### 2.1 Data Collection Methods for Magnetic Adhesion State in Wall-Climbing Robots

In the research of magnetic adhesion wall-climbing robots, sensor technology has always been a critical factor in ensuring their stability and safety. In earlier studies, pressure sensors were widely used to monitor the adhesion force between the robot and the wall surface [24-25]. However, the magnetic adhesion force significantly decreases as the distance between the magnet and the ferromagnetic metal wall increases, making it difficult to effectively install pressure sensors on the magnetic units of the magnetic adhesion wall-climbing robots [20]. Furthermore, although vacuum sensors and optical sensors have also been applied to monitor the robot's adhesion state and ensure stable attachment [26], the use of vacuum sensors is limited by their dependence on the wall material, while optical sensors perform poorly in environments with strong light, uneven lighting, or multiple obstacles, restricting their application in complex environments.

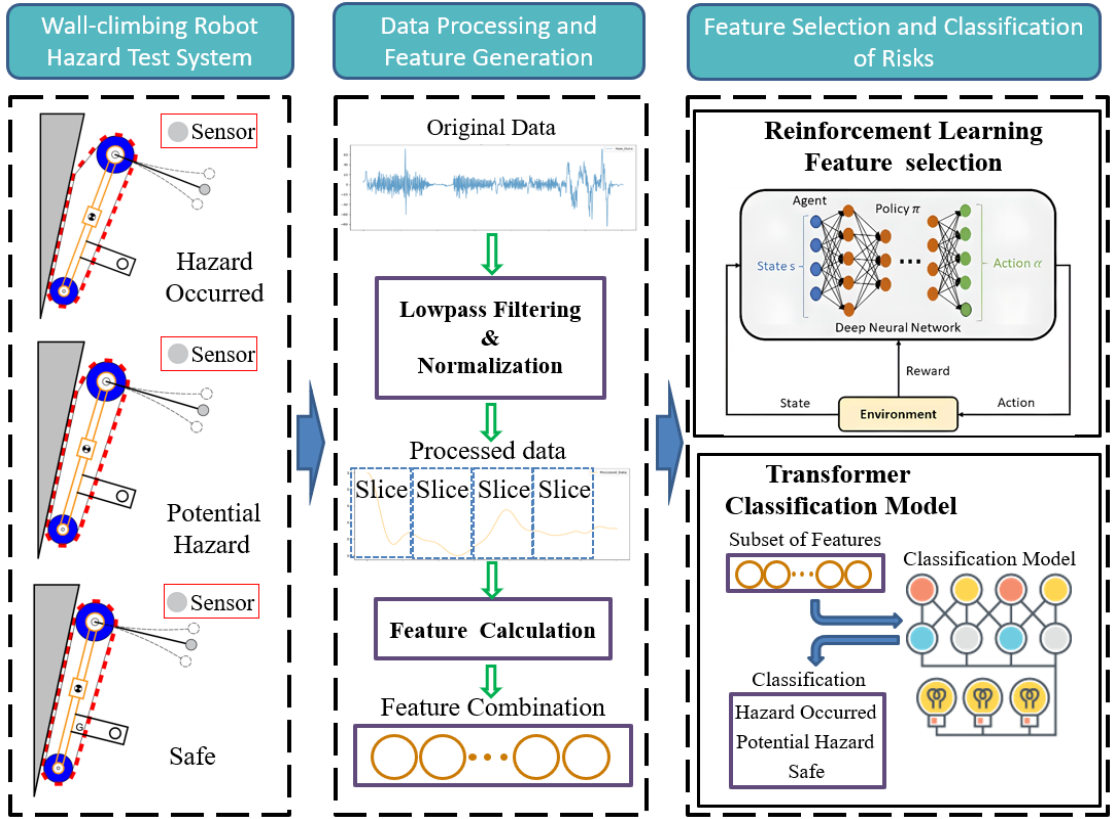

Fig. 1. Research framework of the proposed framework

Vision sensors have been used as an alternative for hazardous state monitoring [27]. However, their accuracy and robustness are significantly reduced in dimly lit environments with many obstructions. Therefore, although many different types of sensors have been used in existing studies to monitor the robot's attachment state, most methods rely on complex and high-cost sensor systems. Research focused on low-cost, simple, and efficient magnetic adhesion monitoring and hazardous state assessment remains relatively scarce. In particular, in the field of magnetic adhesion wall-climbing robots, no mature or reliable solutions have been developed for real-time detection of the magnet's adhesion state.

Micro-Electromechanical Systems (MEMS) sensors, known for their high sensitivity, low cost, small size, and ease of integration, have achieved significant applications across various fields. For example, MEMS attitude sensors are widely used in pipeline robots to assess the robot's positioning accuracy and attitude changes, thereby enhancing its stability and safety in complex environments [28]. In the field of 3D printing, MEMS sensors have enabled real-time diagnostics of 13 common faults by monitoring operational conditions, significantly improving system maintenance efficiency and operational stability [28]. Additionally, MEMS technology is also extensively applied in the automotive industry for precise vehicle yaw estimation and dead reckoning, successfully enabling low-cost driver assistance systems [29].

Despite the successful applications of MEMS sensors in various fields, the use of MEMS technology for detecting the attachment state of magnetic adhesion wall-climbing robots is still in its early stages, requiring further research and exploration.

### 2.2 Application of Reinforcement Learning in Feature Selection

Reinforcement Learning (RL), as an adaptive technology that dynamically adjusts decision-making strategies, has gained widespread attention in the field of feature selection in recent years [30–31]. Traditional feature selection methods, such as filter methods, wrapper methods, and embedded methods, typically rely on static datasets and fixed evaluation criteria, which are unable to effectively handle dynamic environments and complex data distributions. Through interactions with the environment, reinforcement learning can actively explore and learn environmental features, enabling it to quickly adapt to new data tasks in unfamiliar and challenging environments [32].

Existing studies have shown that reinforcement learning can effectively search for the optimal feature subset in high-dimensional feature spaces [23]. For instance, a feedback-based deep reinforcement learning feature selection method has been proposed, which introduces transition similarity measures and leverages deep reinforcement learning to ensure continuous exploration of the state space. This method achieved excellent classification performance when evaluated on nine standard benchmark datasets, demonstrating the potential of deep reinforcement learning in feature selection [33]. Additionally, a new wrapper feature selection method based on a deep artificial curiosity framework has been introduced. This method employs intrinsic reward reinforcement learning with Long Short-Term Memory (LSTM) units, capable of handling feature interaction issues and improving the accuracy of learning models on both synthetic and real-world datasets [34].

Compared to traditional feature selection methods, reinforcement learning has demonstrated stronger capabilities in handling complex nonlinear features and data imbalance issues [35]. This makes reinforcement learning an ideal method for feature selection in this study, where wall-climbing robots operate under various conditions. However, in the research related to perception of wall-climbing robot states, reinforcement learning has rarely been applied for feature selection of perception data, and its use remains in the early stages.

### 3. Methodology

This section analyzes the adhesion mechanism of the wall-climbing robot and describes the attitude data collection strategy, followed by a detailed description of the proposed reinforcement learning-based feature selection model. Finally, the overall process of adhesion state recognition for the wall-climbing robot is summarized in the last subsection.

### 3.1 Magnetic Adhesion Mechanism and Contact Stiffness Analysis of the Wall-Climbing Robot

In operational conditions with high loads or large wall inclination angles, the magnetic adhesion tracked wall-climbing robot may experience the gradual detachment of the track from the wall. As shown in Fig.2, during the climbing process, the track undergoes deformation due to the weight load and wall inclination. When the magnetic pads roll with the track into the upcoming adhesion zone, the distance Lh between the magnet and the wall increases as the tilt angle θ increases, significantly reducing the magnetic attraction. In this scenario, the restoring force from the track deformation is insufficient to maintain normal attachment. For clearer analysis, the magnetic pad lift-off of the wall-climbing robot in this study is transformed into an analysis of the adhesive force exerted by the magnetic pads.

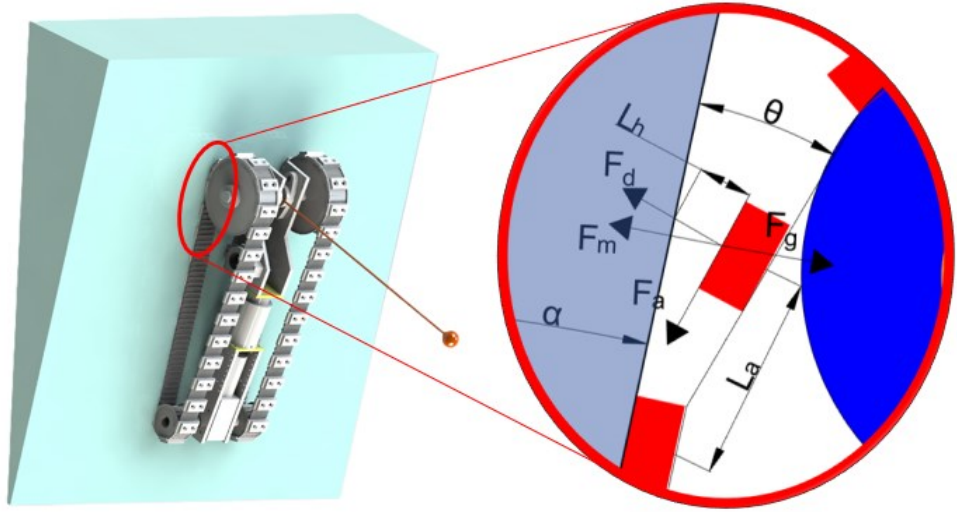

Fig.2. Adhesion Force Analysis of Magnetic Pads on the Wall-Climbing Robot

When the robot climbs upward, as the track rolls into the magnetic adhesion zone ahead, the magnetic attraction force $Fm$ on the magnetic pad is the sum of the magnetic force, track restoring force, and tension force components, while $Fg$ represents the resultant force acting in the opposite direction due to gravity and load. To simplify the description of the interaction of these forces, they are represented using a matrix.

$$F_m = \begin{bmatrix} \dfrac{k}{L_h^2} \\ F_d \cos\theta \\ F_a \cos\theta \end{bmatrix}$$

$$F_g = \left[ (G_a + G_b)\sin(\theta + \alpha) \right]$$

Where k is the magnetic force coefficient, Lh is the distance between the magnetic pad and the wall, Fd and Fa represent the track's restoring force and tension force, respectively, θ is the bending angle of the track, Ga and Gb represent the robot and load's gravitational forces, and α is the wall inclination angle. It can be observed that when ||Fm||≥||Fg||, the magnetic pad is attached, while when ||Fm|| < ||Fg||, the magnetic pad cannot adhere. In cases where attachment is not possible, the robot will experience a gradual reduction in the number of adhered magnetic pads, ultimately leading to a fall.

The connection between the wall-climbing robot and the wall is formed by the rigid attachment of each magnetic pad. Suppose that at a certain moment, the number of adhered magnetic pads is N. The system's total contact stiffness $k_{total}$, natural frequency $\omega_{nat}$, and damping ratio ζ can be expressed as follows:

$$k_{total} = N \cdot k_i$$

$$\omega_{nat} = \sqrt{\frac{k}{m}} = \sqrt{\frac{N \cdot k_i}{m}}$$

$$\zeta = \frac{c}{2\sqrt{mk}} = \frac{c}{2\sqrt{mNk_i}}$$

As shown in the above formula, as N increases or decreases, the contact stiffness of the robot system will change, which in turn affects the changes in the natural frequency $\omega_{nat}$ and damping ratio $\zeta$. Variations in the natural frequency lead to changes in the system's vibration frequency, and the vibration characteristics also affect the robot's adhesion performance with the wall. Therefore, under different numbers of adhered magnetic pads, the system's response to vibration will vary, which in turn influences the adhesion performance of the robot.

### 3.2 Attitude Data Collection

As a tool for measuring three-dimensional motion posture, MEMS attitude sensors are one of the effective means of obtaining system vibration data [36]. In this study, the sensor is connected to the robot body via a carbon fiber rod to collect vibration acceleration data. The introduction of the carbon fiber rod helps amplify the robot's body vibration signals, thus improving the sensitivity of the system's response data. However, changes in the stiffness k of the carbon fiber rod can affect the signal synchronization and amplification effect.

Specifically, when the stiffness of the rod decreases, the gain ‖ ‖ increases, leading to higher sensor sensitivity, but this may result in a loss of signal frequency characteristics. On the other hand, when the stiffness increases, the original characteristics of the signal are better preserved, but the amplification effect weakens, and the sensor sensitivity decreases.

$$|H(\omega)| = \frac{\sqrt{k^2 + (\omega c)^2}}{\sqrt{(k - M\omega^2)^2 + (\omega c)^2}}$$

$$\phi(\omega) = tan^{-1}(\frac{\omega c}{k}k) - tan^{-1}(\frac{\omega c}{k - M\omega^2})$$

It can be observed that when the rod stiffness k decreases, the gain ‖ ‖ increases, which amplifies the vibration signal on the robot body and improves sensor sensitivity. However, this also leads to a loss of signal frequency characteristics. On the other hand, when the stiffness k increases, the original characteristics of the signal are better preserved, but the amplification effect decreases, leading to a decline in sensor sensitivity. Thus, under different rod lengths (which correspond to different stiffness values k), the data characteristics of the sensor on the carbon fiber rod will change in varying degrees, with some conditions offering better performance than others.

It is important to note that, after calibration, such sensors will still display a 1g gravitational acceleration. Additionally, due to inconsistencies in the initial installation direction of the sensors, the variability of the training data increases, which in turn reduces the model's generalization performance. To mitigate the effects of sensor installation direction and gravitational direction changes on model training and prediction performance, this study computes the magnitude of the three-axis acceleration vector $\mathbf{a} = [a_x, a_y, a_z]^T$ output by the sensor. By calculating the magnitude $\|\ \| \overline{a_y^2 + a_z^2}$, the data is transformed into scalar values as input features, thus avoiding data errors caused by inconsistencies in the installation direction. The resulting input signal can be expressed as:

$$F_{input} = \begin{bmatrix} \|\ \| \\ \|\ \| \\ \|\ \| \end{bmatrix}$$

Where $\|\ \| \|\ \| \|\ \| \|$ is the acceleration magnitude calculated by the sensor, and $F_{input}$ is the input feature matrix for the model.

The attitude sensor, vibration rod, and robot body are rigidly connected to ensure there is no relative movement during robot operation. Since the magnetic pads can only fail to adhere during the robot's upward climbing process, the sensor is mounted at the front of the robot. With the help of the attitude sensor, a single-channel vibration signal can be obtained. These signals are collected under different hazardous working conditions and are used to build an intelligent hazardous state classification model.

### 3.3 PPO-Based Reinforcement Learning Feature Selection Model

To select appropriate features for the adhesion state of the wall-climbing robot from MEMS attitude data, reinforcement learning-based feature selection is an effective method for handling high-dimensional data. It can reduce feature redundancy, improve the model's generalization ability, and enhance computational efficiency. This paper proposes a reinforcement learning-based framework (shown in Fig.3), which optimizes the selection of feature subsets progressively through interaction with a specific environment. The Proximal Policy Optimization (PPO) algorithm is adopted, combined with a custom feature selection environment, to model the dynamic characteristics and constraints of feature selection, allowing for dynamic adaptation to the wall-climbing robot's feature requirements.

#### 3.3.1 Environment Design

In this study, we custom-designed a feature selection environment based on the Gymnasium framework and modeled the feature selection problem as a Markov Decision Process (MDP). Let F be the set of all features, and F′⊂F be a feature subset. The state space is S, which is the power set of F, i.e., S=P(F), where P(F) represents all possible feature subsets. Each action represents selecting an unchosen feature f∈F\F′ from the current feature subset, and transitioning the current state to the next state, i.e., the new feature subset F′=F∪{f}. In each iteration, the agent starts with a randomly initialized feature subset and progressively advances by selecting new features until all features are chosen. Through multiple iterations, the agent will find the optimal policy in the state space, thereby selecting the best feature subset.

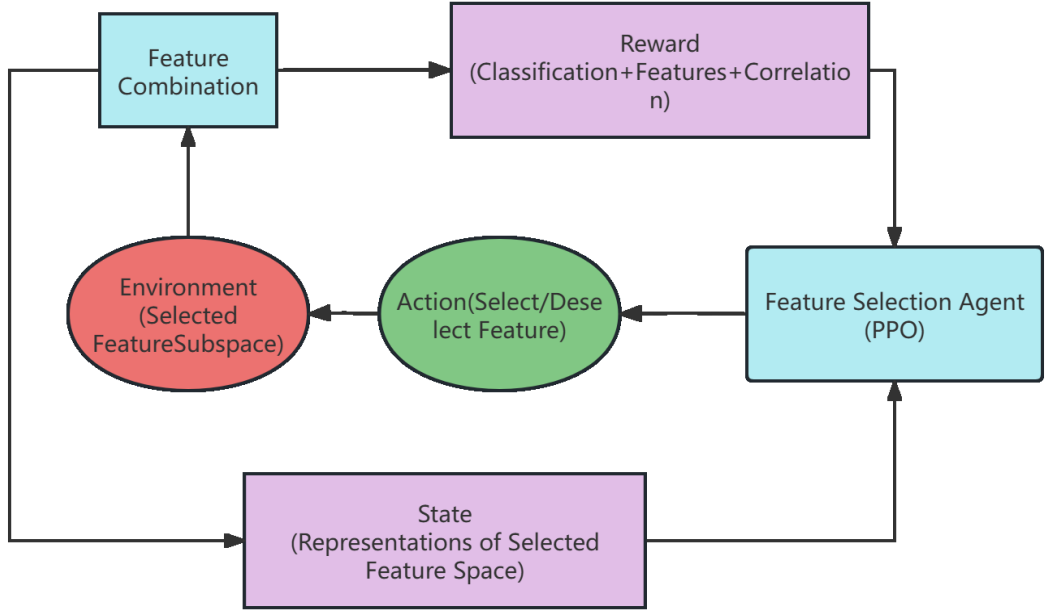

Fig.3. Feature selection framework based on reinforcement learning

Since our goal is to maximize generalization accuracy and minimize the error of the hypothesis learned from the training set, it can be shown that the optimal policy $\pi^*$ is based on the learned hypothesis A, which, in our case, is the output of a random forest classifier.

Let R(F) represent the generalization error of the classifier on the feature subset F. Our objective is to find an optimal feature selection policy $\pi^*$ through reinforcement learning that minimizes the error corresponding to the final selected feature subset. Thus, we aim to minimize:

$$\pi^*(s_t) = \arg\min_{f \in F \backslash F_t} R(F_t \cup \{f\})$$

where $s_t$ represents the state at time step $t$, $F_t$ is the currently selected feature subset, and $R(F_t)$ is the generalization error on the current feature subset.

### 3.3.2 Reward Mechanism

The design of the reward function considers the difference between the current state and the new state after taking an action. The final score of a feature is evaluated based on the average reward obtained by the feature across multiple iterations. The reward function comprehensively accounts for classification performance, the number of features, and the correlation between features, and is defined by the following components:

(1) Classification Performance Reward: The performance on the test set using a random forest classifier is used to evaluate the reward, including common performance metrics such as accuracy, precision, recall, and F1 score. Specifically, the classification performance reward $R_{classification}$ can be defined as:

$$R_{classification} = w_1 \cdot Accuracy + w_2 \cdot Precision + w_3 \cdot Recall + w_4 \cdot F1$$

where $w_1, w_2, w_3, w_4$ is the weight of the corresponding metric.

(2) Feature Count Penalty: The number of features is an important factor affecting model complexity. Too many features not only increase computational cost but may also lead to overfitting. The feature count penalty can be defined as::

$$R_{features} = -\lambda \cdot n$$

where n represents the number of selected features, $\lambda$ is a hyperparameter that controls the penalty strength, and the penalty increases as the number of features grows.

(3) Correlation Penalty: High correlation between features may lead to redundant information and affect the model's generalization ability. To address this, the average correlation of the selected feature subset is calculated, and a penalty is applied based on the correlation. The correlation penalty can be defined as:

$$R_{correlation} = -\alpha \cdot \frac{1}{n(n-1)} \sum_{i=j} corr(x_i, x_j)$$

where $corr(x_i, x_j)$ is the Pearson correlation coefficient between features $x_i$ and $x_j$, and $\alpha$ is a hyperparameter that adjusts the strength of the correlation penalty.

(4) Final Reward: The final reward function R is a weighted combination of the three components above. It can be expressed as:

$$R = \lambda_1 \cdot R_{classification} + \lambda_2 \cdot R_{features} + \lambda_3 \cdot R_{correlation}$$

where $\lambda_1, \lambda_2, \lambda_3$ are the weights assigned to each of the components.

### 3.3.3 Key Hyperparameter Settings

The learning rate, policy update range, and discount factor are essential parameters in model training, and they play a crucial role in the final feature selection performance of the model.

(1) Learning Rate: The learning rate determines the speed of optimization of the objective function and influences whether and when the model converges to a local minimum. An appropriate learning rate enables the objective function to converge within a reasonable time frame. Based on previous training experiences, this study selects an initial learning rate of 0.0001 and increases the learning rate by a factor of 10 after each training iteration to balance convergence speed and stability. To validate this, the loss function and accuracy curves during the training process were recorded, as shown in Fig.4.

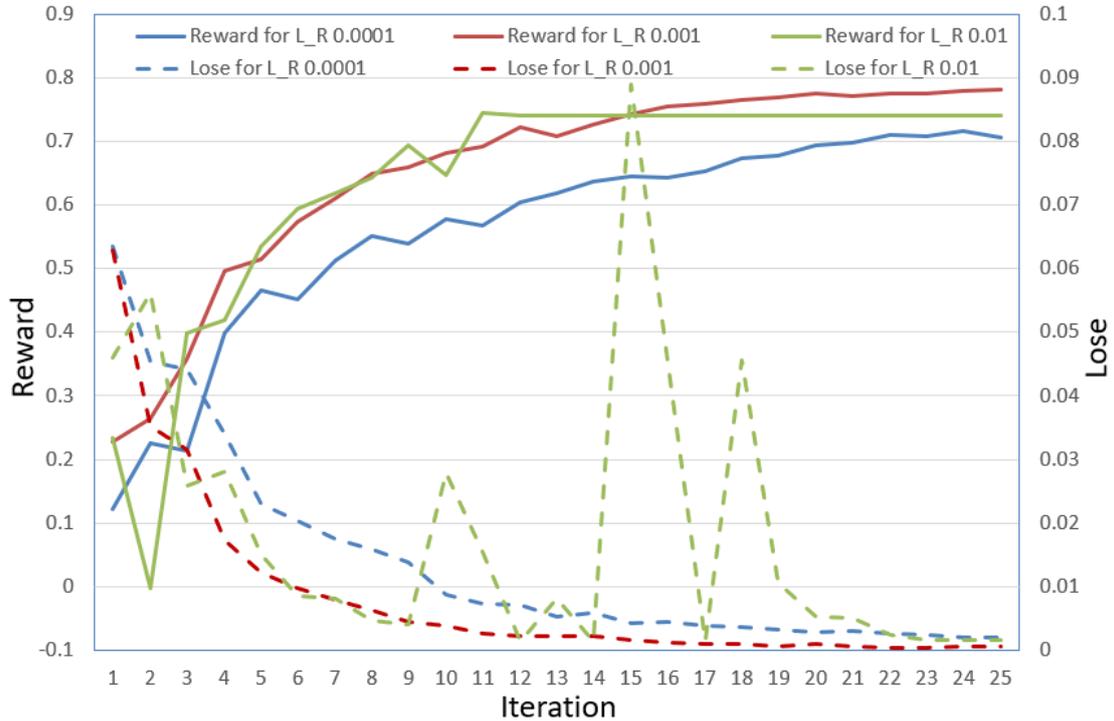

Fig. 4. Comparison of Average Reward and Average Loss of Reinforcement Learning Model at Different Learning Rates

From Fig.4, it can be observed that after approximately 15 training rounds, the reward values stabilize and remain high when the learning rates are 0.0001 and 0.001. However, when the learning rate is 0.01, the reward value is lower. At a learning rate of 0.001, the loss value is consistently lower than at the other two learning rates, with smaller fluctuations. Therefore, considering both the average reward and loss, the final learning rate of 0.001 is selected for model training.

（2）Policy Update: To control the magnitude of each update, this study uses a clipping range to limit the change in actions during each policy update within a reasonable range, thus preventing excessive changes that could lead to policy instability. Based on previous training experiences, this study selects an initial clipping range of clip_range = 0.2 and increases the clipping range by 0.1 after each training iteration. The loss function and accuracy curves during the training process are also recorded, as shown in Fig.5.

（3）Discount Factor: The discount factor controls the weight of future rewards, allowing the agent to consider long-term benefits when selecting features, while avoiding excessive focus on short-term rewards. In this study, the discount factor γ=0.95 is chosen to balance both short-term and long-term rewards.

### 3.4 Wall-Climbing Robot Safety State Recognition Process

To evaluate the effectiveness of the proposed method, which uses signals from MEMS attitude sensors for hazardous state detection of the wall-climbing robot, the overall process of constructing the hazard state classification model is shown in Fig.6, with the specific steps outlined as follows:

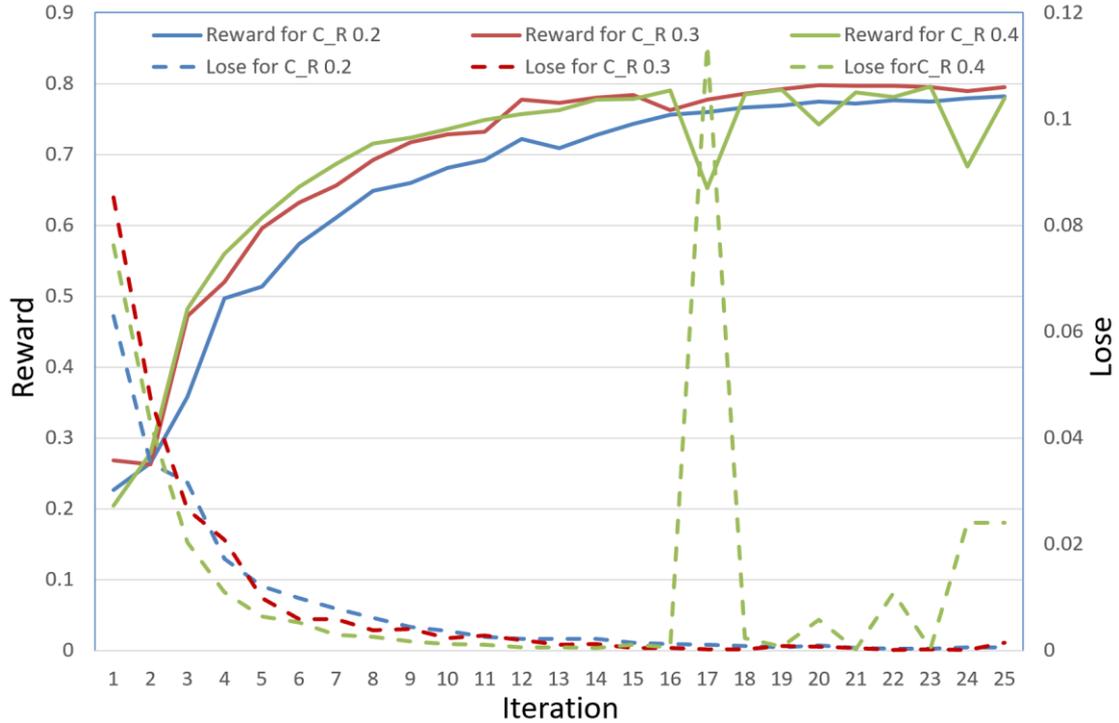

Fig.5. Comparison of Average Reward and Average Loss of Reinforcement Learning Model at Different Clipping Ranges

**Step 1: Data Preprocessing**. the input features are normalized, and low-frequency filtering is applied to remove unnecessary noise, ensuring the quality of the input data. Data preprocessing is crucial for the effective training of the subsequent model.

**Step 2: Feature Selection**. the reinforcement learning framework is used to select key informational features from the high-dimensional feature combinations of the robot's vibration signals. The reinforcement learning model can automatically learn to select the critical feature combinations from high-dimensional data, helping to improve the performance of the classification model.

**Step 3: Classification**. deep learning techniques are used to classify the input features as time series data. The accuracy of label predictions on the test dataset is then evaluated.

4. Experimental Data Preparation

4.1 Raw Data Collection and Labeling

To verify the effectiveness of the proposed hazard perception and classification method for the magnetic adhesion tracked wall-climbing robot, this study designed and built a multi-condition wall-climbing robot falling test system for data collection and testing. As shown in Fig.7, the experimental system consists of the tracked robot, attitude sensor, angle-adjustable steel plate, and load module.

The experiments were conducted in an indoor environment, with the steel plate of the experimental setup fixed by a sturdy support to ensure reliability and repeatability. The experimental operators were trained to skillfully operate the robot, ensuring the accuracy of data collection. The designed load for the wall-climbing robot is 5 kg, and a fixed load of 5 kg was used in the experiments. The experimental variables were adjusted by setting the angle between the steel plate and the vertical plane. The specific experimental conditions included two angles: 55° and 65°, which represent the robot's limit working conditions.

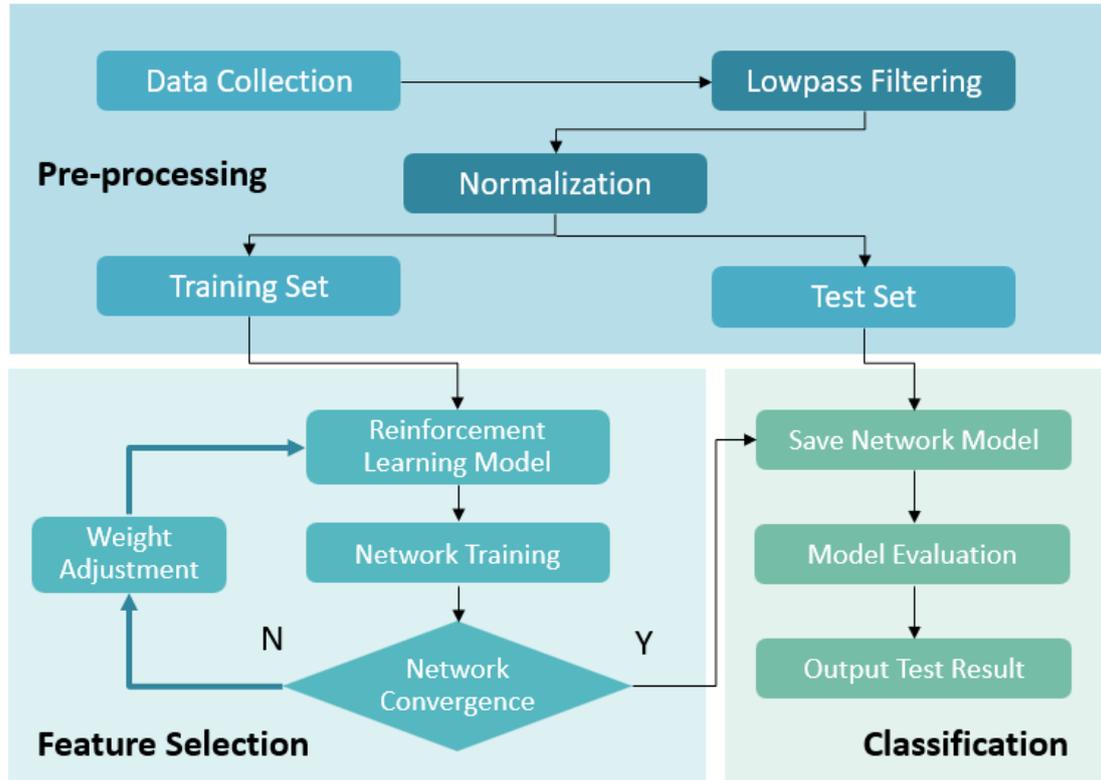

Fig.6. Hazard Classification Flowchart. The hazardous state detection process is divided into three parts: data preprocessing, feature selection, and result classification.

To collect motion data, the robot was connected to the MEMS attitude sensor via a carbon fiber rod. The carbon fiber rod has a length of 40 cm and a diameter of 3 mm, effectively amplifying the vibration signals during the robot's operation. The MEMS attitude sensor is connected via 5G communication, with the following specifications:

Table 1: Key Parameters of the WT9011DCL-BT50 Attitude Sensor

| No. | Parameter Name | Value |
| --- | --- | --- |
| 1 | Measurement Range | ±2000°/s |
| 2 | Sampling Frequency | 0.2-100 kHz |
| 3 | Resolution | 0.061 (°/s)/LSB |
| 4 | Static Zero Bias | ±0.5~1°/s |
| 5 | Temperature Drift | ±0.005~0.015 (°/s)/°C |
| 3 | Sensitivity | ≤0.015°/s rms |

The sampling frequency of the attitude sensor was set to 100 Hz. During the experiment, the robot moved along the steel plate's trajectory at a constant speed of 0.02 m/s, collecting three-axis (X, Y, Z) angular velocity signals. Data collection started from the initial state when all six magnetic pads were attached, and ended when the number of magnetic pads decreased to four. To ensure labeling accuracy, signals from magnetic pads that were partially separated or in partial contact were not collected. Each run collected 50,000 sample points. Each experimental angle was repeated three times, collecting a total of 300,000 sample points, with 70% of the data used for model training and 30% for model testing.

During data collection, when the first magnetic pad at the top rolls into the magnetic adhesion area but cannot adhere, and the last magnetic pad at the bottom rolls out of the adhesion surface, it is labeled as "Potential Hazard." When two magnetic pads fail to adhere, it is labeled as "Hazardous State," and data collection is stopped. The safety state labels are defined as shown in Table 2.

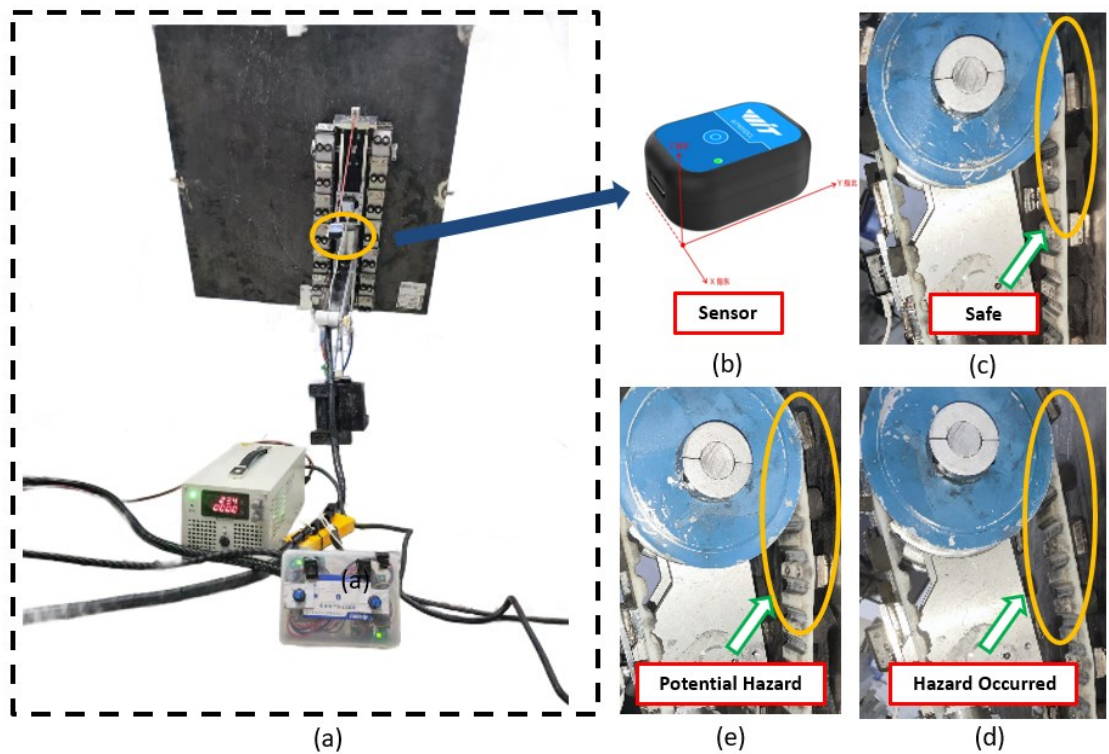

Fig.7. Collecting Attitude Data from the Wall-Climbing Robot. (a) Test setup system;(b) MEMS attitude sensor;(c) Potential hazardous state;(d) Hazardous state occurrence.

Table 2: Definition and Corresponding Labels for Wall-Climbing Robot Hazardous States

| No. | Hazardous State | Label |
|---|---|---|
| 1 | 6 Magnetic Pads Attached | Safe |
| 2 | 5 Magnetic Pads Attached | Potential Hazard |
| 3 | 4 Magnetic Pads Attached | Hazard Occurred |

**4.2 Feature Generation and Data Pre-analysis**

Based on the raw dataset collected in Section 4.1, data preprocessing and feature generation were performed. During the data preprocessing phase, the raw data was first subjected to low-frequency filtering and normalization. The continuous signals were then segmented into slices of 32 data points each, and the time-domain and frequency-domain features of each segment were calculated. A total of 17 features were generated. To further validate the model, 6 noise features were added to the real features, resulting in a 23-dimensional feature vector. Table 3 shows the generated features and noise categories.

Table 3: Class Names and Categories for All Feature Dimensions

| Feature Dimension | Feature Type | Category |
|---|---|---|
| 1 | Mean | Time Domain |
| 2 | Standard Deviation | Time Domain |
| 3 | Maximum Value | Time Domain |
| 4 | Minimum Value | Time Domain |
| 5 | Norm | Time Domain |
| 6 | Energy | Time Domain |
| 7 | Kurtosis | Time Domain |
| 8 | Skewness | Time Domain |
| 9 | Simple Mean Absolute Value | Time Domain |
| 10 | Autocorrelation | Time Domain |
| 11 | Autocorrelation Lag 2 | Time Domain |
| 12 | Autocorrelation Lag 3 | Time Domain |
| 13 | Mean Power Frequency | Frequency Domain |
| 14 | Median Frequency | Frequency Domain |
| 15 | Total Power | Frequency Domain |
| 16 | Maximum Power Spectral Density | Frequency Domain |
| 17 | Zero Crossing Rate | Frequency Domain |
| 18 | Random Noise 1 | Synthetic/Noise Features |
| 19 | Random Noise 2 | Synthetic/Noise Features |
| 20 | Feature Perturbation Noise 1 | Synthetic/Noise Features |
| 21 | Feature Perturbation Noise 2 | Synthetic/Noise Features |
| 22 | Temporal Perturbation Noise 1 | Synthetic/Noise Features |
| 23 | Temporal Perturbation Noise 2 | Synthetic/Noise Features |

From Table 3, it can be seen that the real features consist of statistics, spectral characteristics, and autocorrelation features computed from signal slices, while the noise features are added to simulate uncertainty and measurement errors, thereby improving the model's robustness in real-world environments.

## 5. Experimental Results and Discussion

### 5.1 Experimental Setup

The wall-climbing robot and the testing platform used in this study are prototypes designed and manufactured by the laboratory and have been put into use, ensuring the reproducibility of the experiments. All code was implemented using the TensorFlow framework and ran on a computing platform equipped with an NVIDIA 3060 12GB GPU.

### 5.2 Training Analysis of the Reinforcement Learning Model

The training process of the reinforcement learning model exhibits significant dynamic characteristics, especially in the changes of key training metrics such as average episode reward, value function loss, feature selection quantity, and feature selection similarity. As the training progresses, the changes in these metrics reflect the model's gradual optimization of its strategy and improvement in execution efficiency.

(1) Average Episode Reward and Value Function Loss

During the model's training process, the average episode reward significantly increases, while the loss consistently decreases. This indicates that the model is gradually improving its strategy during the learning process and is able to make decisions that align more closely with the task objectives. Higher episode rewards and lower loss values reflect that the agent is receiving more positive feedback during task execution, suggesting that the model is effectively selecting and optimizing the feature selection strategy. By continually adjusting its strategy, the model can better adapt to environmental feedback and achieve higher rewards during task execution.

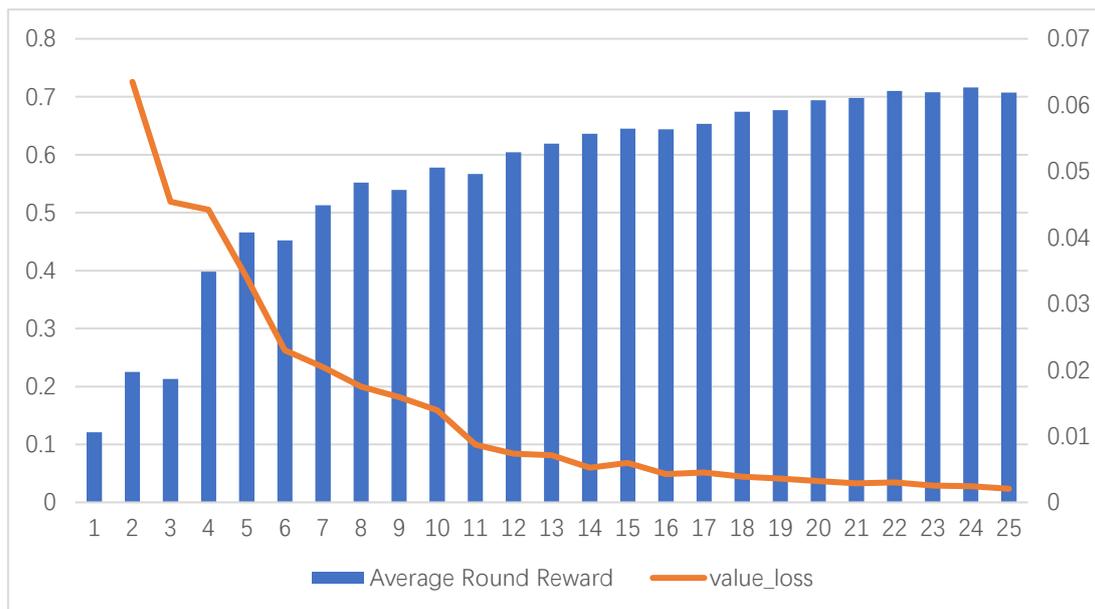

Fig. 8. Average Episode Reward and Loss during Training

(2) Training Stability Evaluation

To assess the stability of the feature selection process, we recorded the number of features selected by the reinforcement learning model in each round of training. A total of 8,595 rounds of training were performed, with every 50 rounds grouped together as a slice. We then calculated the average number of features selected per slice. Additionally, we computed the Jaccard similarity between different episodes. By calculating the similarity between feature subsets selected across consecutive episodes, we can analyze whether the model consistently selects the same feature set across multiple rounds. The statistical results are shown in Fig. 9.

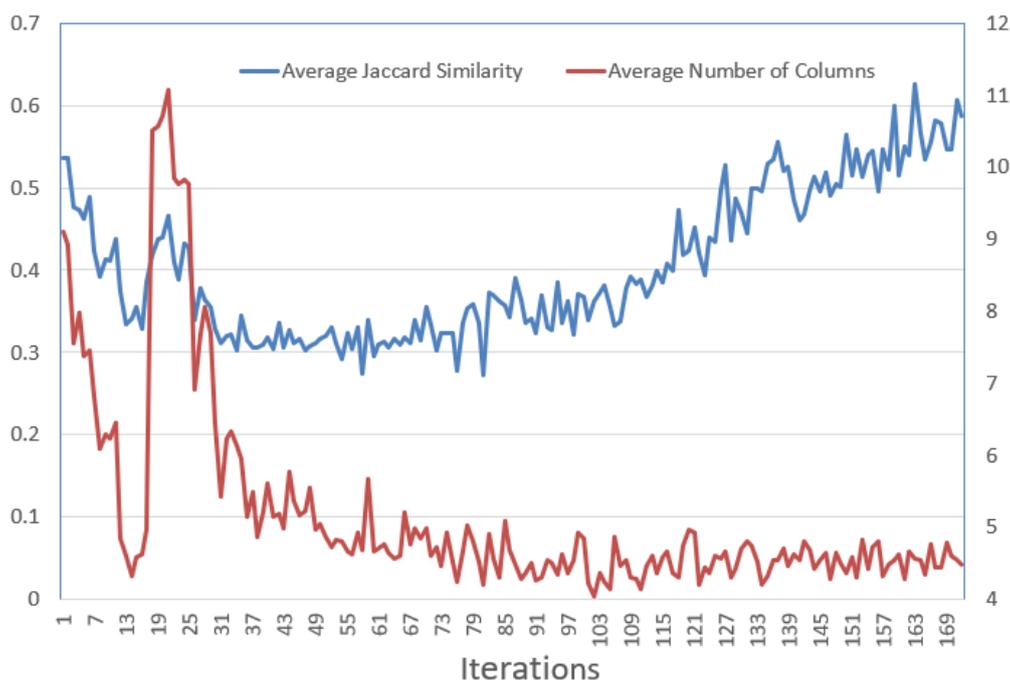

Fig.9. Feature Selection Stability and Jaccard Similarity during Training

As shown, the number of features selected gradually decreases as training progresses and tends to stabilize. In the experiments, we observed that as training progressed, the Jaccard similarity increased and stabilized, indicating that the model is continuously optimizing and converging to a relatively stable feature selection strategy.

(3) Training Redundancy Evaluation

Additionally, to evaluate redundancy, we selected every 5 rounds of training as a slice and chose the round with the maximum number of selected features within each slice. We then calculated the correlation matrix between the features to ensure low redundancy among the selected features. Using a heatmap, as shown in Fig.10, we can visually display the correlation between the features and further validate the quality of feature selection.

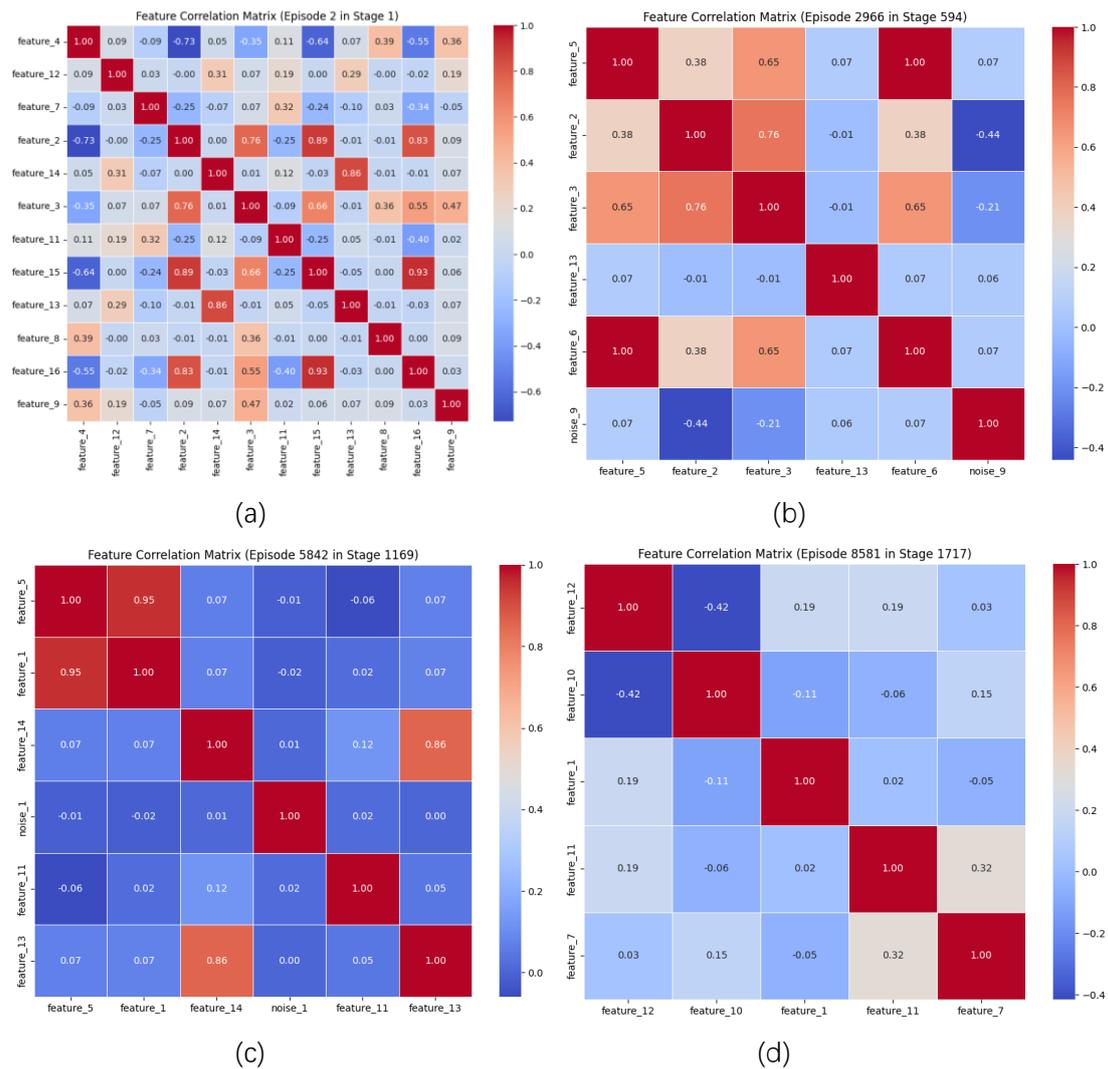

Fig.10. Feature Selection Correlation Matrix Heatmap. (a) Selection at the beginning of the model training;(b) Selection at the mid-point of model training;(c) Selection in the later stages of model training;(d) Selection at the end of model training.

By analyzing the results, it is evident that the model progressively optimized its strategy during training, leading to improved task execution efficiency and increased rewards. In the early stages, the model's decision-making process was more random, resulting in longer episode lengths and lower rewards. However, as training progressed, the model gradually improved its decision-making ability by continuously optimizing its strategy, leading to increased efficiency, reduced episode steps, and enhanced rewards, demonstrating more mature decision-making capabilities.

### 5.3 Comparison with Typical Feature Selection Methods

To further demonstrate the effectiveness of the proposed method, three typical feature selection methods were compared:

Univariate Statistical Selection (SKB): This method calculates the linear correlation between features and the target variable using the ANOVA F-value statistic, and employs a forward selection strategy to retain the top NN features with the highest significance level.

Recursive Feature Elimination (RFE): This method uses a random forest classifier (with a default of 100 trees) as the base learner and iteratively eliminates the features with the smallest feature weights until the predefined feature dimension is reached.

Tree Model Feature Importance Ranking (RFFI): This method directly trains a random forest classifier and calculates the feature importance score based on the decrease in the Gini index. The subset of features with the top n importance scores is selected.

For the wall-climbing robot's vibration state characteristics, we used CNN-LSTM as the classification model for hazardous state detection [37].

To facilitate a better comparison, we used a comprehensive classification method evaluation metric, F-score, defined as:

$$F-score = \frac{2TP}{2TP+FP+FN}$$

Where TP, FP, and FN represent the number of true positive samples, false positive samples, and false negative samples, respectively. The value of F-score ranges between 0 and 1, with a higher F-score indicating better classification performance.

First, to obtain comparative results across different operating conditions, training samples from four different conditions were combined. The different conditions here refer to the sensor data obtained from the wall-climbing robot under different loads. The average F-score values for the four combinations are listed in Table 4.

Table 4: Comparison of Average F-score for Different Feature Selection Methods Across Different Operating Condition Combinations

| Method | 1kg | 1kg+2kg | 1kg+2kg+3kg | 1kg+2kg+3kg+5kg |
|---|---|---|---|---|
| SKB | 0.8741 | 0.849 | 0.7814 | 0.7022 |
| RFE | 0.9315 | 0.8732 | 0.8025 | 0.7204 |
| RFFI | 0.9381 | 0.885 | 0.8214 | 0.6418 |
| PPO | 0.9448 | 0.9206 | 0.9396 | 0.9106 |

As shown in Table 4, with the increasing combination of condition data, the average F-score values for each method generally decrease. However, PPO consistently outperforms the other three comparison methods, with a weaker reduction in F-score. Particularly, when data from all four condition combinations are mixed, PPO's average F-score is 0.1902 higher than the second-best method, RFE. This indicates that PPO is highly suitable for dynamic operating conditions.

We also performed an analysis of the classification accuracy for the labels. Fig.11 presents the F-score for the three safety labels across different reinforcement learning methods. The results show that the F-score for all three labels based on the PPO method are consistently higher than those of the other three methods. Additionally, it is evident that the F-score for PPO's three label classifications have smaller variances, making PPO more stable compared to the other methods.

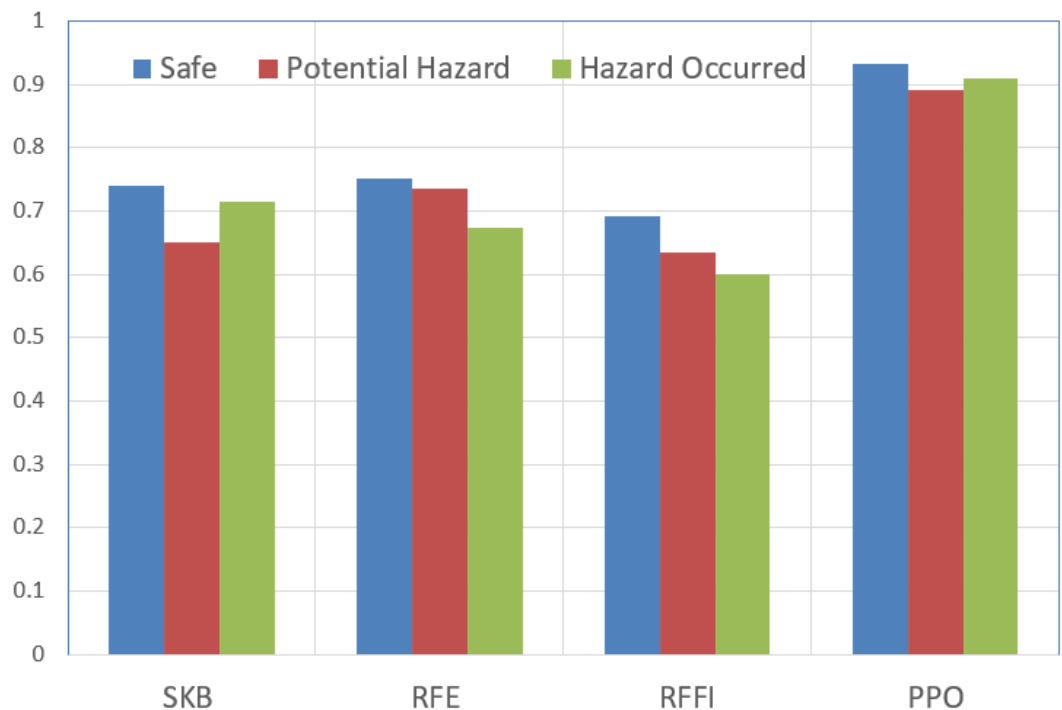

Fig.11. F-score values of three labels under different feature selection methods

To further validate the superiority of PPO over other methods, we assessed its ability to adapt to noise level variations in the environment, thereby studying PPO's generalization capability. This was also compared with SKB, RFE, and RFFI. Since the wall-climbing robot used in this study operates in an environment where interference is more likely to occur due to sudden changes in vibration signal time and amplitude, salt-and-pepper noise is more appropriate for this scenario. We introduced three levels of salt-and-pepper noise conditions: (1) Low-density noise (1% probability for salt and pepper noise each, total 2%); (2)Medium-density noise (3% probability for salt and pepper noise each, total 6%); (3)High-density noise (5% probability for salt and pepper noise each, total 10%).

We used the original vibration signals for training and then applied the above-mentioned noise. Fig. 12 plots the test accuracy of the four methods across a range of salt-and-pepper noise from 0% to 10%. The results show that the performance of all methods decreases as the noise level increases. However, the test accuracy of PPO consistently remains higher than the other three methods, with a smaller decline in accuracy, indicating that our method exhibits better generalization capability to environmental changes.

### 6. Results and Discussion

This study focuses on the safety state perception of tracked wall-climbing robots, designing an effective and easy-to-implement data collection strategy, feature selection, and classification model. Through the analysis of the wall-climbing robot's lift-off process, we introduced a simple yet effective posture data collection strategy. This strategy uses a carbon fiber vibration rod with an attitude sensor mounted at the end to monitor the robot's adhesion posture to the wall.

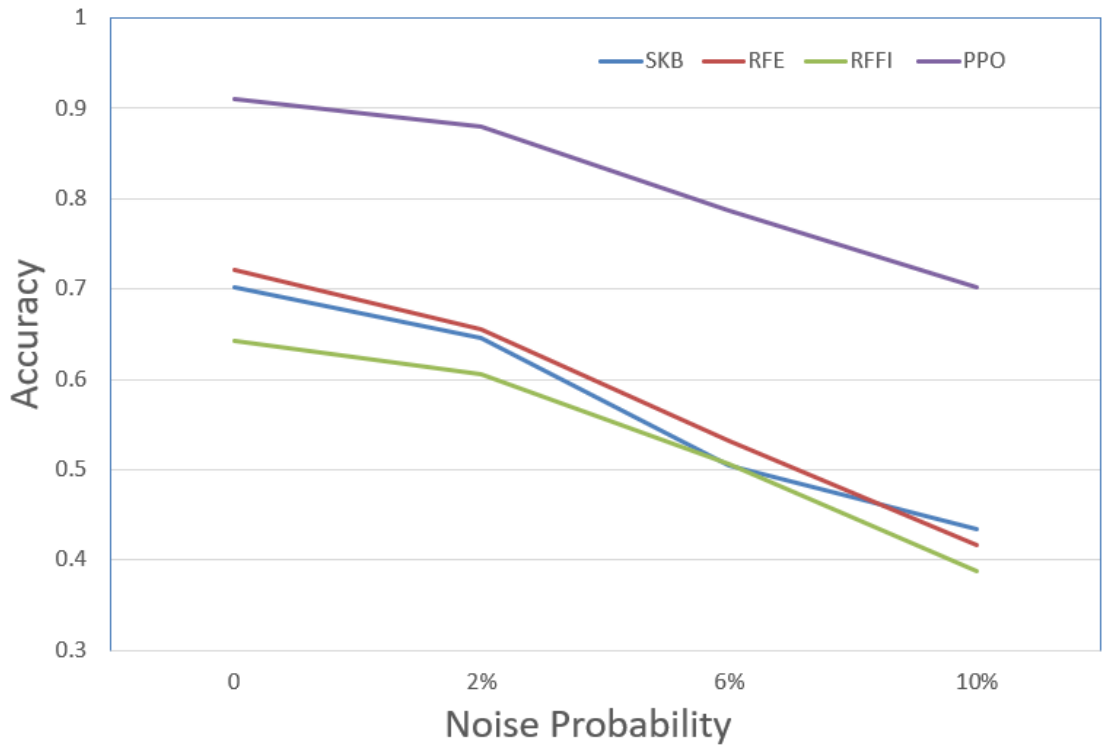

Fig. 11. Test Accuracy Rate of the Model under Different Noise Addition Ratios

To extract features from the attitude data, we proposed a reinforcement learning feature selection method that combines the PPO algorithm with a custom dynamic feature selection environment, enabling cross-condition application for the robot. We established a test platform that includes a wall-climbing robot and an adjustable angle wall to create different recognition tasks.

We first conducted comparison experiments, and the results demonstrate that the proposed feature selection strategy achieves good performance in hazardous state recognition for wall-climbing robots. The strategy outperformed classical feature selection methods in both single-condition and multi-condition classification results.

Additionally, in this study, the training and testing data came from the same wall-climbing robot and test platform. However, handling training and testing data from different wall-climbing robots and test platforms is a more challenging research task. We will focus on these tasks in future work.